\documentclass[preprint,12pt]{elsarticle}

\usepackage{amsmath,amsfonts}
\usepackage{algorithmic}
\usepackage{array}
\usepackage[tight,footnotesize]{subfigure}
\usepackage{textcomp}
\usepackage{stfloats}
\usepackage{url}
\usepackage{verbatim}
\usepackage{graphicx}
\def\BibTeX{{\rm B\kern-.05em{\sc i\kern-.025em b}\kern-.08em
    T\kern-.1667em\lower.7ex\hbox{E}\kern-.125emX}}
\usepackage{balance}
\usepackage{booktabs}
\usepackage{caption}
\usepackage{makecell}
\usepackage{adjustbox}
\usepackage{wrapfig}
\usepackage[linesnumbered,ruled]{algorithm2e}
\usepackage{amssymb}
\usepackage{lineno}
\usepackage{float}
\usepackage{algorithmic}
\usepackage[tight,footnotesize]{subfigure}
\usepackage{booktabs}
\usepackage{caption}
\usepackage{makecell}
\usepackage{adjustbox}
\usepackage{wrapfig}
\usepackage{xcolor}
\usepackage[linesnumbered,ruled]{algorithm2e}

\newcommand{\cmark}{\ding{51}}%
\newcommand{\xmark}{\ding{55}}%

\journal{Journal Name}

\begin{document}
\sloppy
\setlength{\parskip}{0pt}

\begin{frontmatter}

\title{ASTRIDE: A Security Threat Modeling Platform for Agentic-AI Applications}

\author[label1]{Eranga Bandara}
\ead{cmedawer@odu.edu}

\author[label2]{Amin Hass}
\ead{amin.hassanzadeh@accenture.com}

\author[label1]{Ross Gore}
\ead{rgore@odu.edu}

\author[label1]{Sachin Shetty}
\ead{sshetty@odu.edu}

\author[label1]{Ravi Mukkamala}
\ead{mukka@odu.edu}

\author[label1]{Safdar H. Bouk}
\ead{sbouk@odu.edu}

\author[label3]{Xueping Liang}
\ead{xuliang@fiu.edu}

\author[label4]{Ng Wee Keong}
\ead{awkng@ntu.edu.sg}

\author[label5]{Kasun De Zoysa}
\ead{kasun@ucsc.cmb.ac.lk}

\author[label6]{Aruna Withanage}
\ead{aruna@effectz.ai}

\author[label6]{Nilaan Loganathan}
\ead{nilaan@effectz.ai}

\address[label1]{Old Dominion University, Norfolk, VA, USA}
\address[label2]{Accenture Technology Labs, Arlington, VA, USA}
\address[label3]{Florida International University, USA}
\address[label4]{Nanyang Technological University, Singapore}
\address[label5]{University of Colombo, Sri Lanka}
\address[label6]{Effectz.AI}

\begin{abstract}

AI agent-based systems are becoming increasingly integral to modern software architectures, enabling autonomous decision-making, dynamic task execution, and multimodal interactions through large language models (LLMs). However, these systems introduce novel and evolving security challenges, including prompt injection attacks, context poisoning, model manipulation, and opaque agent-to-agent communication, that are not effectively captured by traditional threat modeling frameworks. In this paper, we introduce ASTRIDE, an automated threat modeling platform purpose-built for AI agent-based systems. ASTRIDE extends the classical STRIDE framework by introducing a new threat category, \textit{A for AI Agent-Specific Attacks}, which encompasses emerging vulnerabilities such as prompt injection, unsafe tool invocation, and reasoning subversion, unique to agent-based applications. To automate threat modeling, ASTRIDE combines a consortium of fine-tuned vision-language models (VLMs) with the OpenAI-gpt-oss reasoning LLM to perform end-to-end analysis directly from visual agent architecture diagrams, such as data flow diagrams(DFDs). LLM agents orchestrate the end-to-end threat modeling automation process by coordinating interactions between the VLM consortium and the reasoning LLM. Our evaluations demonstrate that ASTRIDE provides accurate, scalable, and explainable threat modeling for next-generation intelligent systems. To the best of our knowledge, ASTRIDE is the first framework to both extend STRIDE with AI-specific threats and integrate fine-tuned VLMs with a reasoning LLM to fully automate diagram-driven threat modeling in AI agent-based applications.

\end{abstract}


\end{frontmatter}

\section{Introduction}

The rapid proliferation of AI agent-based systems—autonomous software entities empowered by large language models (LLMs) and capable of dynamic task execution, multimodal perception, and self-directed reasoning—has fundamentally transformed modern application architectures~\cite{agentic-ai, agentsway}. These agents are increasingly embedded in workflows ranging from intelligent assistants to autonomous decision-making systems and multi-agent environments. However, their emerging behaviors and the reliance on complex language-driven interactions introduce novel security challenges that extend beyond the scope of traditional software threats. These include prompt injection attacks, context poisoning, reasoning subversion, and unintended tool invocation, all of which exploit the flexible and opaque nature of agent decision pipelines~\cite{llm-attack}. Existing threat modeling methodologies, including the widely adopted STRIDE framework, are not well-equipped to capture the emerging attack surfaces unique to agentic AI applications~\cite{threat-modeling-stride, deep-stride}. These frameworks rely heavily on manual inspection of system architectures, demand significant domain expertise, and lack native support for AI-specific vulnerabilities such as prompt manipulation, context or memory poisoning, and inter-agent influence. Consequently, critical threats may go undetected until late in the development lifecycle, posing significant risks to the integrity, availability, and trustworthiness of AI-driven systems~\cite{threat-modeling-challenges}.

To address this gap, we introduce ASTRIDE, an automated threat modeling platform specifically designed for AI agent-based applications. ASTRIDE extends the classical STRIDE framework by introducing a new threat category \textit{A for AI Agent–Specific Attacks} which captures emerging security risks unique to agentic workflows. These include instruction manipulation (e.g., prompt injection), unsafe reasoning-driven tool use, and the misuse of agent memory or context windows. ASTRIDE combines a consortium of fine-tuned vision-language models (VLMs)~\cite{vision-language-model} with the OpenAI-gpt-oss reasoning LLM~\cite{reasoning-llms} to perform end-to-end threat analysis directly from visual architecture diagrams, such as data flow diagrams. Each VLM is trained on a custom dataset of AI agent topologies annotated with both traditional STRIDE vectors and AI-specific threats, enabling robust component-level vulnerability detection. The VLM outputs are aggregated and synthesized by OpenAI-gpt-oss, which performs high-level reasoning to generate a coherent and explainable threat model. LLM agents~\cite{agentic-ai} orchestrate the end-to-end threat modeling automation process by coordinating interactions between the VLM consortium and the reasoning LLM. To ensure practical deployment, all models are fine-tuned using the Unsloth library with QLoRA-based quantization, enabling low-latency inference on resource-constrained hardware~\cite{llamafactory-unsloth}.

Our evaluation shows that ASTRIDE provides an accurate, scalable, and interpretable solution for threat modeling in next-generation AI systems. To our knowledge, this is the first framework to both extend STRIDE with AI-specific threats and integrate fine-tuned VLMs with a reasoning LLM to fully automate visual, diagram-driven threat modeling for AI agent-based applications. The following are our main contributions of this research. 

\begin{enumerate}
    \item Propose a novel threat modeling framework for capturing the attack surfaces of agentic AI applications by extending the traditional STRIDE methodology. 
    \item Automate the end-to-end threat modeling process for agentic AI applications using a consortium of fine-tuned vision-language models and the OpenAI-gpt-oss reasoning model. \item Utilize a fine-tuned vision-language model consortium to analyze threat model images (e.g., data flow diagrams) and generate ASTRIDE-based threat predictions for agentic AI applications. 
    \item Incorporate the OpenAI-gpt-oss reasoning model to synthesize and finalize threat assessments based on the outputs of the vision-language model consortium.
\end{enumerate}

The rest of the paper is organized as follows: Section 2 presents related work and positions our approach in the context of existing AI-enabled threat modeling frameworks. Section 3 describes the overall system architecture of the proposed platform. Section 4 outlines the core functionalities and operational workflow. Section 5 provides details on the implementation and performance evaluation. Finally, Section 6 concludes the paper and discusses directions for future research.

\section{Related Work}

Several researchers have explored AI-based approaches to threat modeling, each addressing different aspects of securing intelligent systems. Mauri and Damiani~\cite{stride-ai-ml} extended STRIDE for AI/ML lifecycles, emphasizing tailored security assessments. The ADMIn framework~\cite{admin} by Kumar et al. categorizes threats to AI software into attacks on datasets, models, and inputs. Yang et al.\cite{threatmodeling-llm} proposed using LLMs to automate threat modeling in banking systems, while Mollaeefar et al.\cite{pillar} introduced PILLAR, which integrates LLMs with the LINDDUN framework to automate privacy threat classification. In healthcare, STRIDE was adapted for LLM-based systems to model component-level threats~\cite{threat-modeling-healthcare}. Auspex~\cite{auspex} introduced ‘tradecraft prompting’ to embed threat modeling expertise into generative AI systems. Lastly, STRIDE-AI~\cite{stride-ai} by Zhou et al. offers an asset-centric model for identifying threats in ML components like datasets and training environments. 

Table~\ref{t_bc_platforms} provides a comparative analysis of existing frameworks across key dimensions, including support for fine-tuning on structured diagram datasets, runtime integration of LLMs and VLMs, vision-language threat modeling capabilities, reasoning LLM usage, and modular consortium-based orchestration of multiple specialized models. Most prior work lacks vision-language understanding of system diagrams or treats threat modeling as static text generation, offering limited ability to capture agent-specific attack surfaces such as prompt injection, context poisoning, or unsafe tool invocation addressed in ASTRIDE.

In contrast, our proposed ASTRIDE platform extends STRIDE with an additional category for AI Agent–Specific Attacks and uniquely combines fine-tuned vision-language models and the OpenAI-gpt-oss reasoning LLM to automate threat analysis directly from system diagrams. ASTRIDE supports fine-tuning on structured diagram datasets, ensures interoperability across multiple models, and introduces decision-level synthesis to resolve conflicting outputs. By integrating architectural understanding, multimodal reasoning~, and safeguards against agent-specific risks such as prompt injection and unsafe tool invocation~\cite{llm-attacks, agentic-ai}, ASTRIDE advances the state of the art in automated threat modeling for agentic AI applications.

\begin{table*}[!htb]\centering
\vspace{0.1in}
\caption {AI-enabled Threat Modeling Framework Comparison}
\begin{adjustbox}{width=1\textwidth}
\label{t_bc_platforms}
\begin{tabular}{lcccccc}
\toprule
\thead{Platform} & \thead{Domain} & \thead{Fine-tuning\\Support} & \thead{Running LLM} & \thead{Vision LM\\Support} & \thead{Reasoning LLM\\Support} & \thead{LLM Consortium\\Support} \\
\midrule
Astride & Threat Modeling & \cmark & \makecell{Llama-3, Pixtrel, Qwen2\\OpenAI-gpt-oss} & \cmark & \cmark & \cmark \\
STRIDE for AI/ML~\cite{stride-ai-ml} & AI/ML Systems & \xmark & \xmark & \xmark & \xmark & \xmark \\
ADMIn~\cite{admin} & AI-based Software & \xmark & \xmark & \xmark & \xmark & \xmark \\
ThreatModeling-LLM~\cite{threatmodeling-llm} & Banking Systems & \xmark & GPT-based & \xmark & \xmark & \xmark \\
PILLAR~\cite{pillar} & Privacy (LINDDUN) & \xmark & GPT-4 & \xmark & \xmark & \xmark \\
LLM-based Healthcare STRIDE~\cite{threat-modeling-healthcare} & Healthcare Systems & \xmark & GPT-4 & \xmark & \xmark & \xmark \\
Auspex~\cite{auspex} & General Threat Modeling & \xmark & GPT-4 & \xmark & \xmark & \xmark \\
STRIDE-AI~\cite{stride-ai} & ML System Assets & \xmark & \xmark & \xmark & \xmark & \xmark \\
\bottomrule
\end{tabular}
\end{adjustbox}
\end{table*}

\section{System Architecture}

\begin{figure}[H]
\centering{}
\includegraphics[width=5.2in]{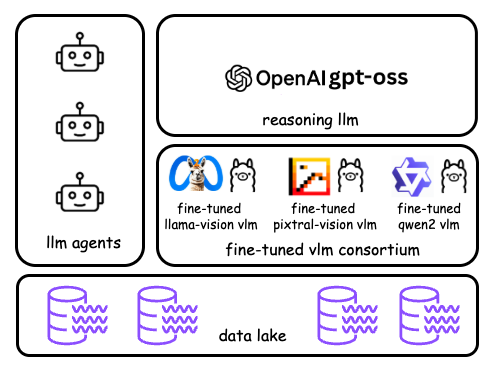}
\DeclareGraphicsExtensions.
\caption{ASTRIDE system architecture.}
\label{indy528-architecture}
\end{figure}

Figure~\ref{indy528-architecture} illustrates the overall architecture of the proposed platform, which is composed of four core components: (1) Data Lake, (2) Vision-Language Models, (3) OpenAI-gpt-oss Reasoning LLM, and (4) LLM Agents. A brief description of each component is provided below.

\subsection{Data Lake}

Data Lake serves as the foundational infrastructure for the ASTRIDE platform, providing centralized storage, management, and access to the diverse and large-scale datasets required for training and fine-tuning VLM and reasoning LLMs for automated threat modeling~\cite{vision-language-model}. Specifically, this layer aggregates a curated corpus of labeled threat modeling diagrams, including Data Flow Diagrams associated with threat vectors and trust boundaries. These datasets are critical for teaching VLMs to understand the structural and semantic patterns in system diagrams and to associate them with potential security threats.

\subsection{Vision-Language Models}

VLMs form the analytical layer of the ASTRIDE platform, enabling it to interpret complex threat modeling diagrams and autonomously extract threat insights. This layer consists of a consortium of fine-tuned VLMs, each trained on a large, labeled dataset of system architecture and data flow diagrams annotated with associated threat vectors, trust boundaries, and component-level metadata~\cite{vision-language-model, nurolense}. Rather than relying on a single model, the ASTRIDE platform employs a multi-model ensemble strategy, where each VLM independently analyzes input diagrams and outputs structured threat observations. These include AI agent-specific attacks, as well as potential STRIDE attacks spoofing, tampering, repudiation, information disclosure, denial of service, and privilege escalation threats - each mapped to specific components and data flows in the diagram. The use of diverse VLMs enhances resilience against individual model biases and ensures broader semantic coverage across a range of threat modeling styles and visual representations.

\subsection{OpenAI-gpt-oss Reasoning LLM}

The OpenAI-gpt-oss Reasoning LLM~\cite{gpt-oss, reasoning-llms} represents the cognitive layer of the ASTRIDE platform, responsible for high-level reasoning, synthesis, and refinement of threat predictions derived from complex system diagrams. In the context of ASTRIDE, this layer performs the final aggregation and decision-making function in the Agentic-AI application threat modeling pipeline. Predictions generated by the VLM consortium, each VLM offering a unique perspective on AI-Agent specific attacks and STRIDE-based attacks(spoofing, tampering, repudiation, information disclosure, denial of service, and elevation of privilege threats), are collected and structured by the LLM Agent Layer into a composite prompt~\cite{llm-attack, vindsec-llams}. This prompt is then submitted to OpenAI-gpt-oss, which uses its reasoning abilities to synthesize a cohesive, prioritized, and contextually validated threat model.

By cross-analyzing the output of multiple VLMs, OpenAI-gpt-oss ensures that conflicting or incomplete assessments are reconciled into a consistent and trustworthy threat modeling representation. It also applies advanced reasoning to contextual cues (e.g., system roles, trust boundaries, and domain-specific risk patterns), enabling deeper insights than purely visual or rule-based systems can achieve~\cite{gpt-oss, o3}.

\begin{figure}[H]
\centering{}
\includegraphics[width=5.2in]{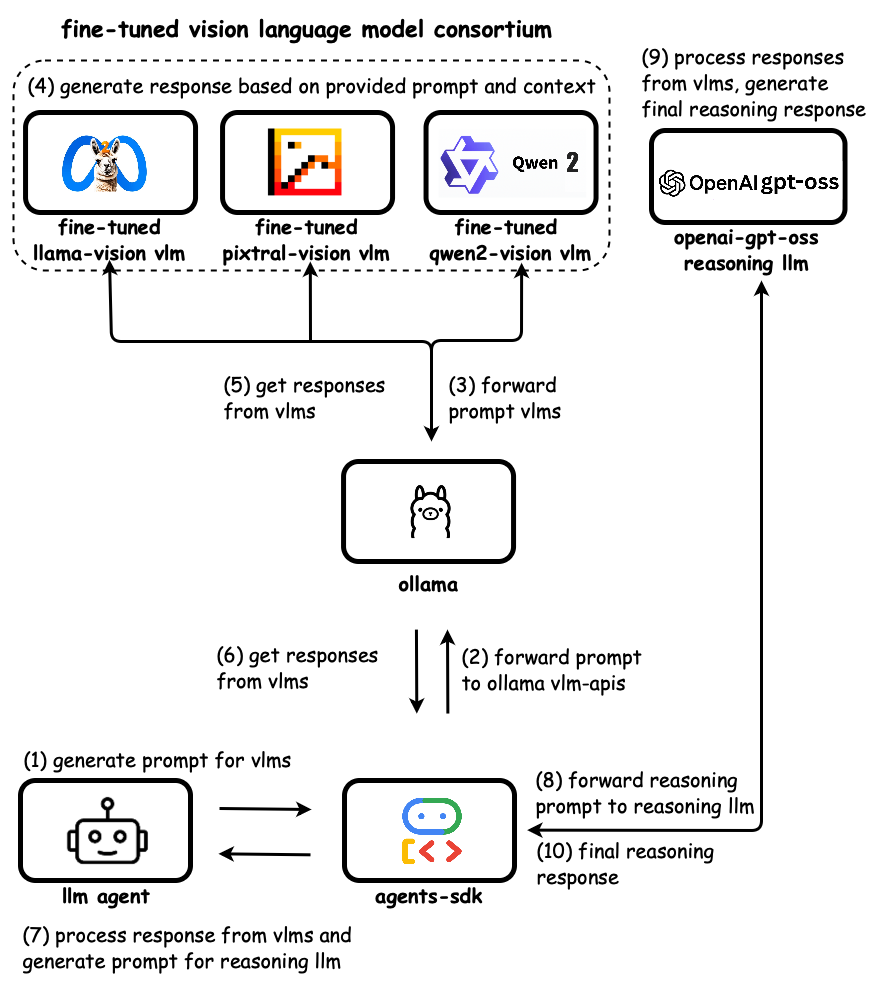}
\DeclareGraphicsExtensions.
\caption{ASTRIDE threat prediction flow}
\label{llama2-flow}
\end{figure}

\subsection{AI/LLM Agents}

AI/LLM Agents~\cite{agentic-ai} serve as the orchestration layer of the ASTRIDE platform, managing coordination between the VLM consortium and the OpenAI-gpt-oss reasoning LLM~\cite{gpt-oss, deep-psychiatric}. The agents generate customized prompts tailored to extract both STRIDE and Agentic-AI-specific threat vectors from the fine-tuned VLMs. Each VLM independently analyzes the input and returns structured threat insights, as illustrated in Figure~\ref{llama2-flow}. Smart contracts then aggregate and reformat the VLM predictions into a unified reasoning prompt for OpenAI-gpt-oss, which synthesizes the output into a coherent and context-sensitive final threat model~\cite{reasoning-llms, mcc}.

\section{Platform Functionality}

There are four main functionalities of the platform: 1) Data Lake Setup, 2) Vision language model Fine-Tuning, 3) Threat prediction by Fine-tuned VLMs, and 4) Final Prediction by OpenAI-gpt-oss LLM. This section goes into the specifics of these functions.

\subsection{Data Lake Setup}

The Data Lake serves as the foundational layer of the ASTRIDE platform, designed to store, organize, and manage the comprehensive datasets required for fine-tuning VLMs on threat modeling tasks. Unlike traditional datasets focused on image classification or natural language, the ASTRIDE Data Lake contains visual system artifacts such as data flow diagrams, system architecture diagrams, and associated textual annotations, including labeled threat vectors, component descriptions, trust boundary definitions, and mitigation factors, which streamlines the fine-tuning process~\cite{mistral-fine-tune, devsec-gpt}.

\subsection{Vision-Language Model Fine-Tuning}

\begin{figure}[H]
\centering{}
\includegraphics[width=5.3in]{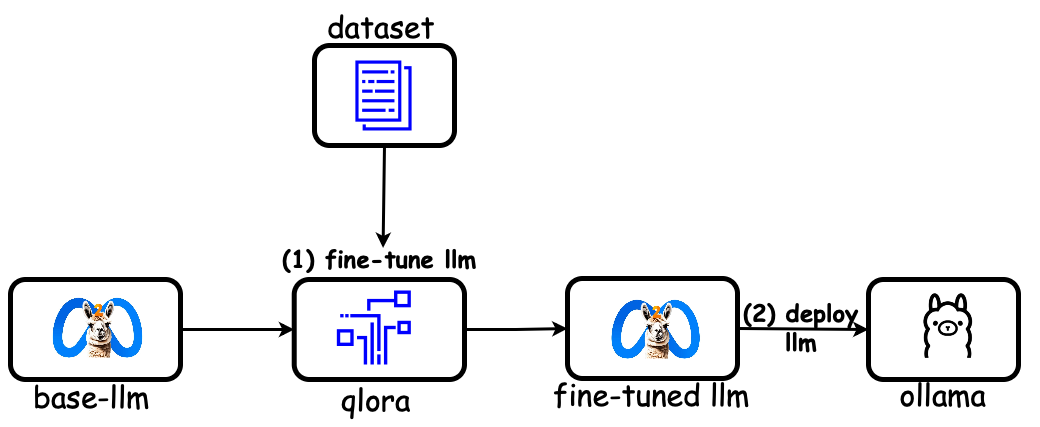}
\DeclareGraphicsExtensions.
\caption{Fine-tune VLMs with Qlora and deploy with Ollama.}
\label{llm-fine-tune}
\end{figure}

The next phase in the ASTRIDE workflow is the fine-tuning of VLMs using the curated, labeled dataset of threat modeling diagrams and threat annotations stored in the Data Lake. This step involves adapting general-purpose VLMs(e.g. \textit{Llama-Vision, Pix2Struct, and Qwen2-VL}~\cite{vistion-language-model-comparison, proof-of-tbi}) to the specific domain of cybersecurity threat modeling. These models are fine-tuned to accurately associate system architecture and data flow diagram features with corresponding threat vectors, trust boundaries, system component roles, and mitigation. Through this fine-tuning pipeline, ASTRIDE transforms general VLMs into domain-specialized threat prediction agents capable of extracting actionable threats from real-world agentic-ai system diagrams with high precision and consistency, as illustrated in Figure~\ref{llm-fine-tune}.

\subsection{Threat Prediction by Fine-Tuned VLMs}

Once the VLMs are fine-tuned using labeled threat modeling data, the next key functionality of the ASTRIDE platform is to extract threat predictions from new system diagrams using the VLM consortium. When a new data flow diagram or system architecture image is submitted, the LLM agent layer orchestrates the interaction with each fine-tuned VLM and initiates a distributed threat analysis process.

To ensure accurate and context-aware predictions, it employs custom prompt engineering, crafting model-specific instructions that inject system metadata, domain context, and task objectives alongside visual input~\cite{}. These prompts instruct each VLM to analyze the image and extract threat vectors(e.g., AI-Agents specific threats and STRIDE threats) and to associate them with relevant system components and data flows~\cite{threat-modeling-stride}.

Each model in the consortium independently processes the input diagram and produces a structured set of predictions, including identified threat types, associated components, severity levels, and recommended mitigations. These outputs are returned to the LLM agent layer, which collects, formats and prepares them for further synthesis using the OpenAI-gpt-oss Reasoning Layer~\cite{reasoning-llms}.

\subsection{Final Prediction and Reasoning by OpenAI-gpt-oss LLM}

To ensure the precision, consistency, and reliability of the threat models generated, ASTRIDE employs a consensus-based reasoning mechanism powered by the OpenAI-gpt-oss LLM. Rather than relying on the output of a single VLM, the platform aggregates threat predictions from multiple independently fine-tuned VLMs. These intermediate outputs, each offering different perspectives on the same visual input, are then synthesized by OpenAI-gpt-oss, which serves as the final reasoning and decision layer within the platform~\cite{reasoning-llms}.

\begin{figure}[H]
\centering{}
\includegraphics[width=5.3in]{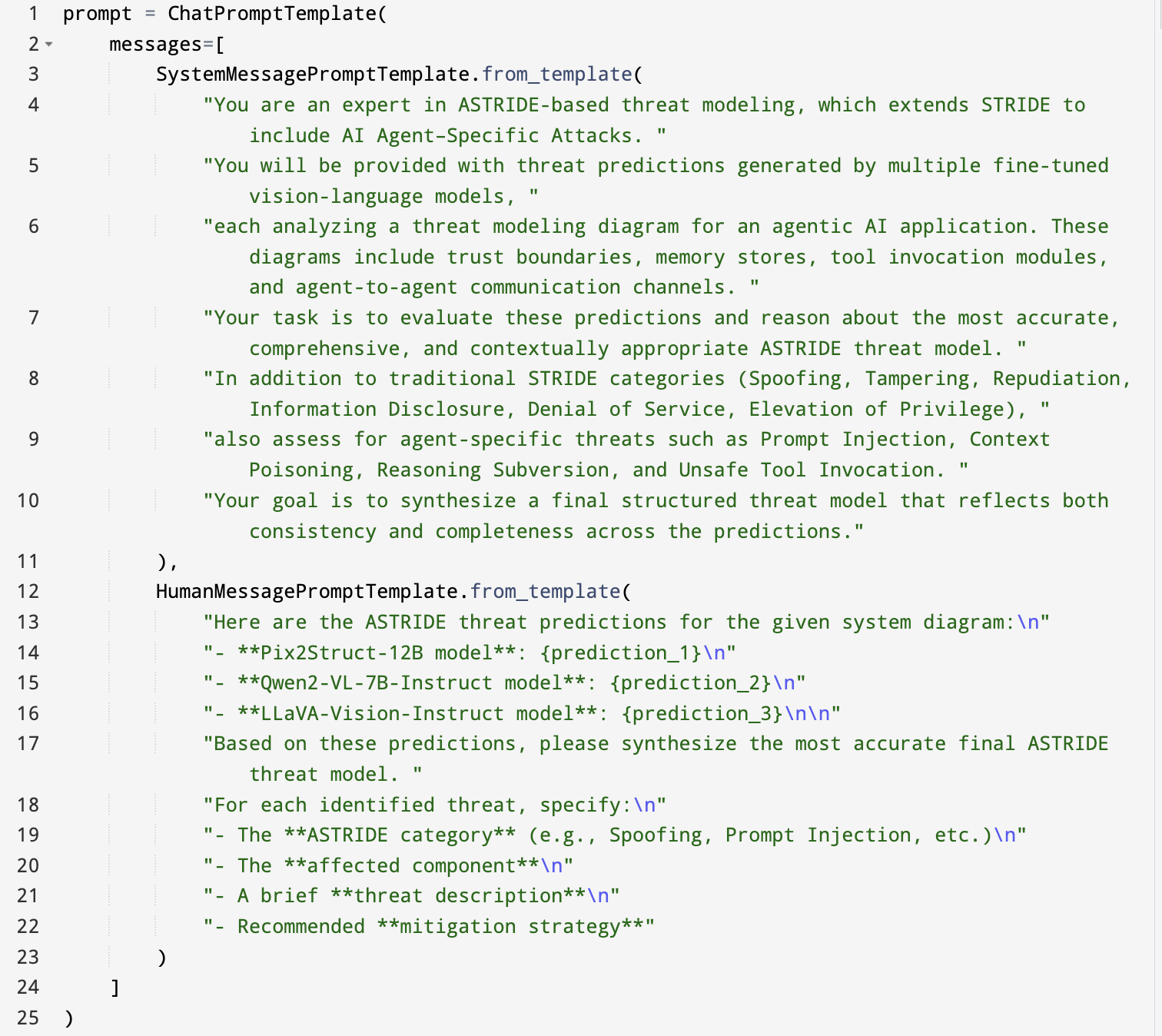}
\DeclareGraphicsExtensions.
\caption{Prompt for OpenAI-gpt-oss reasoning LLM.}
\label{prompt}
\end{figure}

As a reasoning-centric LLM, OpenAI-gpt-oss is uniquely suited to interpret and compare the predictions from different VLMs. Contextual analysis is performed to validate, reconcile, and rank threats, ultimately generating a unified threat model that reflects the most accurate and complete understanding of the system diagram. To guide this reasoning process, LLM agents construct structured prompts that encapsulate each VLM’s output, including identified threat vectors, affected components, and proposed mitigation strategies. An example of this prompt structure is shown in Figure~\ref{prompt}.

\section{Implementation and Evaluation}

A functional prototype of the ASTRIDE platform has been implemented to validate the feasibility and performance of automated threat modeling from visual system representations. The core analytical layer is composed of three fine-tuned VLMs—\textit{Llama-Vision, Pix2Struct, and Qwen2-VL}~\cite{vistion-language-model-comparison}—each trained to interpret architectural diagrams and extract ASTRIDE-based security insights. The fine-tuning process utilized a synthetically generated dataset comprising system architecture and data flow diagrams annotated with trust boundaries, threat vectors, and mitigation strategies. Each diagram in the dataset was paired with a structured JSON representation containing labeled components, identified threats (e.g., AI-Agents specific threats as well as STRIDE threats), and corresponding mitigation techniques. 

Fine-tuning was performed using the \texttt{Unsloth} library on Google Colab, leveraging NVIDIA A100 and Tesla TPU resources to enable efficient training cycles. The dataset consisted of approximately 1,200 annotated records, automatically generated in the form of Mermaid diagrams~\cite{mermaid} representing system architectures (e.g., data flow diagrams, component diagrams, and trust-boundary layouts). Each entry was structured in a conversational instruction-tuning format, as required by Unsloth~\cite{llamafactory-unsloth}, and included a \texttt{content} field (containing the Mermaid diagram prompt), a \texttt{type} field (describing the diagram category), and an \texttt{instruction} field guiding the model to identify STRIDE threats and recommend corresponding mitigations. The dataset was split into 2/3 for training, 1/6 for validation, and 1/6 for testing. Training was completed in approximately 1,627 seconds (27.12 minutes). The peak reserved memory during training was 14.605 GB, with actual training consumption reaching 5.853 GB, representing 39.69\% of total memory usage and 99.03\% of peak allocation. These metrics demonstrate that structured threat modeling using visual input and LLM fine-tuning can be accomplished efficiently, even on moderate-scale datasets. Post fine-tuning, all models were optimized using \texttt{QLoRA}~\cite{qlora}, allowing quantized deployment on edge or resource-constrained hardware. These VLMs were then hosted on the \texttt{Ollama} framework, which provides an efficient runtime environment for inference~\cite{ollama}. The reasoning layer of the proposed ASTRIDE platform is implemented using the OpenAI-gpt-oss LLM. The LLM agent layer is implemented with the OpenAI-Agents-SDK~\cite{agentic-ai}. Platform performance is evaluated in two key areas: Evaluation of fine-tuned VLMs and Evaluation of OpenAI-gpt-oss Reasoning LLM.

\subsection{Evaluation of Fine-tuned VLMs} 

\begin{figure}[H]
\centering{}
\includegraphics[width=5.3in]{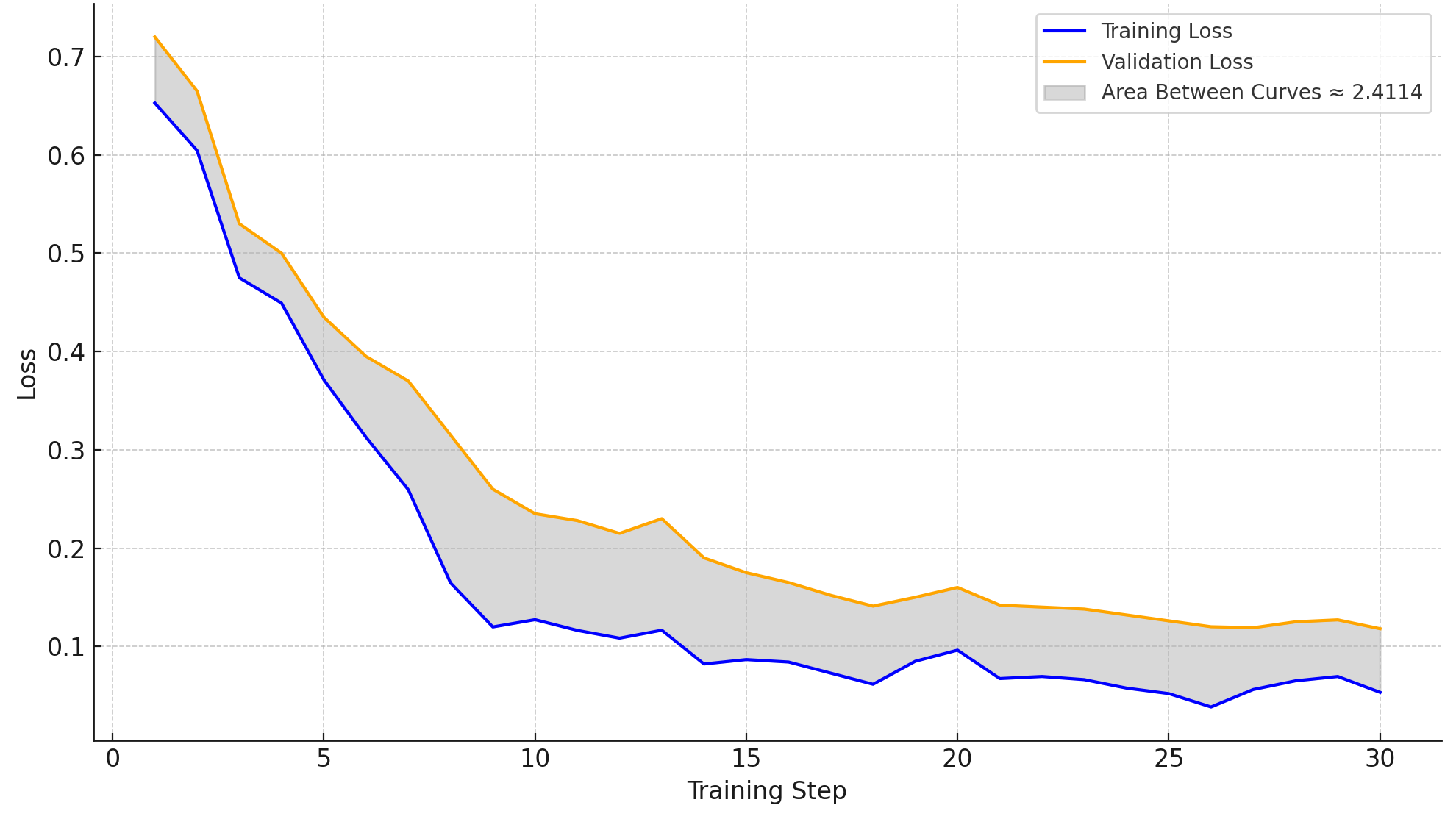}
\DeclareGraphicsExtensions.
\caption{Training loss and validation loss during fine-tuning of the Llama-3.2-11B-Vision-Instruct VLM.}
\label{unsloth-tranning-validation-loss}
\end{figure}

\begin{figure}[H]
\centering{}
\includegraphics[width=5.3in]{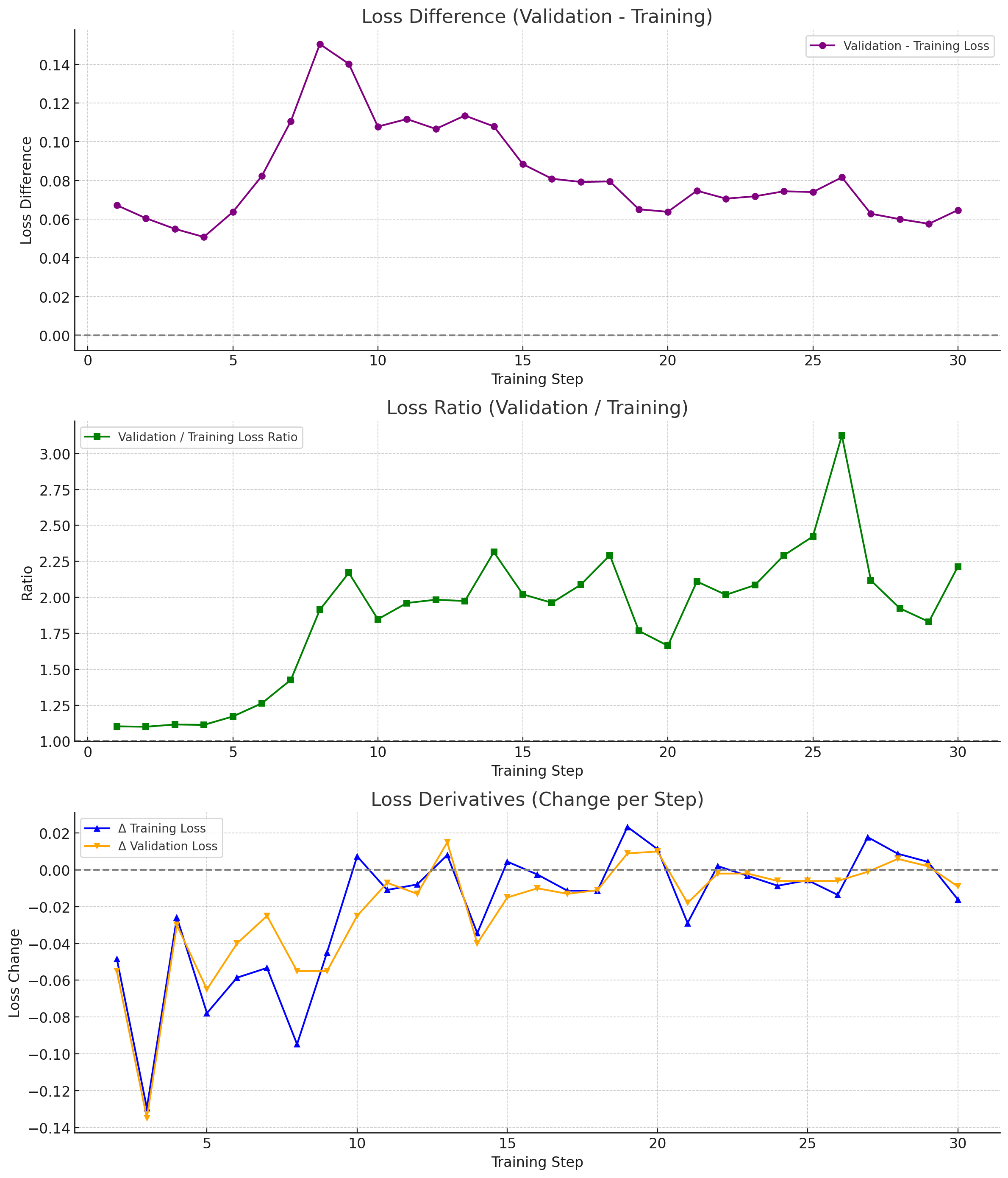}
\DeclareGraphicsExtensions.
\caption{Ratio of training to validation loss during the fine-tuning of the Llama-3.2-11B-Vision-Instruct VLM.}
\label{unsloth-loss-ratio}
\end{figure}

In this evaluation, we first assessed the training and validation loss during the fine-tuning process of the VLMs for diagram-based threat modeling. These metrics, visualized in Figure~\ref{unsloth-tranning-validation-loss}, demonstrate the models’ progressive learning across training steps. Furthermore, Figure~\ref{unsloth-loss-ratio} captures multiple key training dynamics, including the loss difference, loss ratio, and loss derivatives over training steps, offering valuable insights into the model’s convergence behavior and generalization performance. The consistently positive loss difference (validation loss exceeding training loss) suggests signs of overfitting, especially at steps with noticeable spikes. The loss ratio, ranging from 1.0 to 3.0, highlights varying degrees of generalization, where a lower ratio reflects better alignment between training and validation performance. Additionally, the loss derivatives reveal rapid initial improvements followed by smaller, oscillating changes, indicating stabilization or saturation in the learning process. 

\begin{figure}[H]
\centering{}
\includegraphics[width=5.3in]{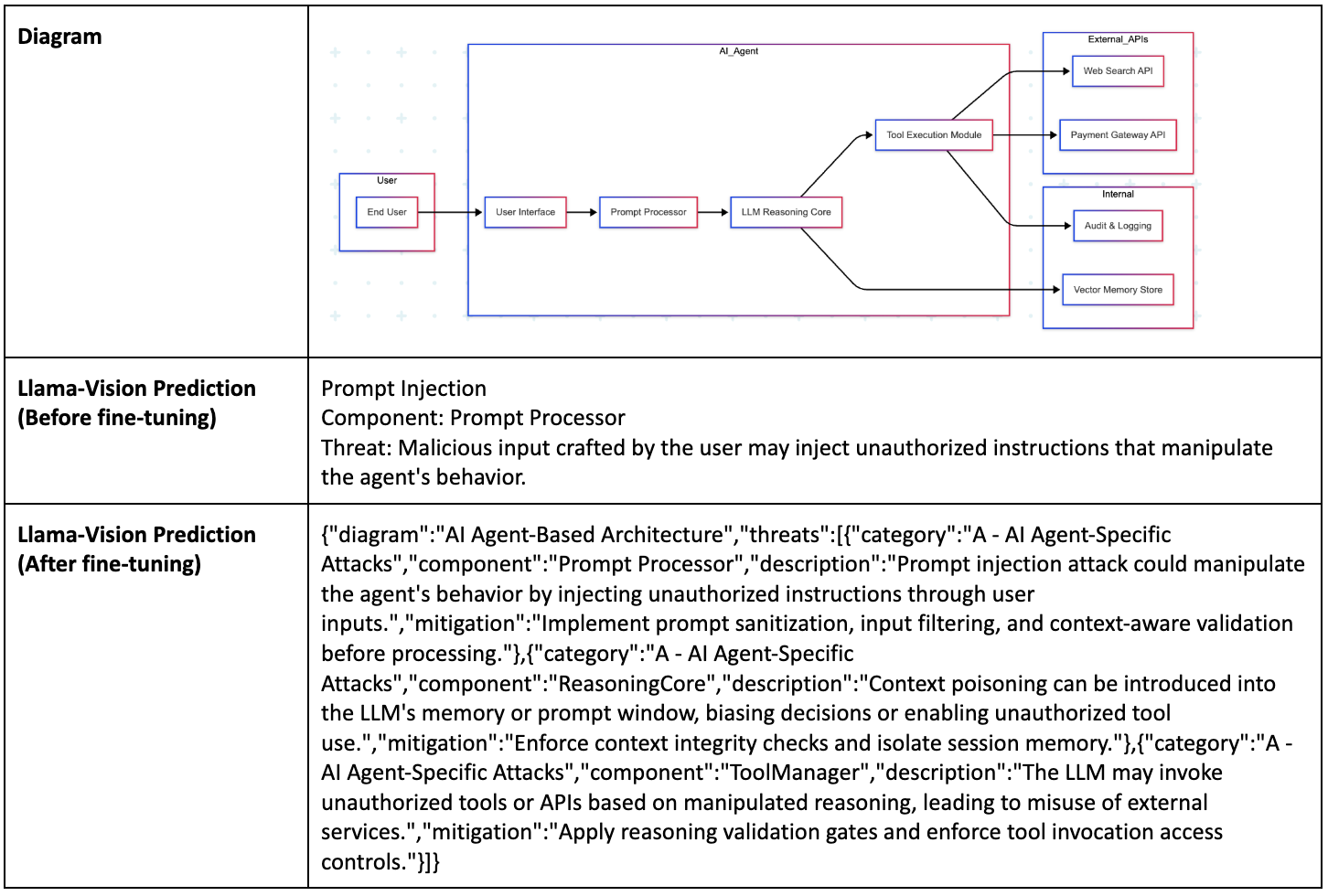}
\DeclareGraphicsExtensions.
\caption{The prediction results of Llama-3.2-11B-Vision-Instruct VLM.}
\label{llama-vistion-result}
\end{figure}

\begin{figure}[H]
\centering{}
\includegraphics[width=5.3in]{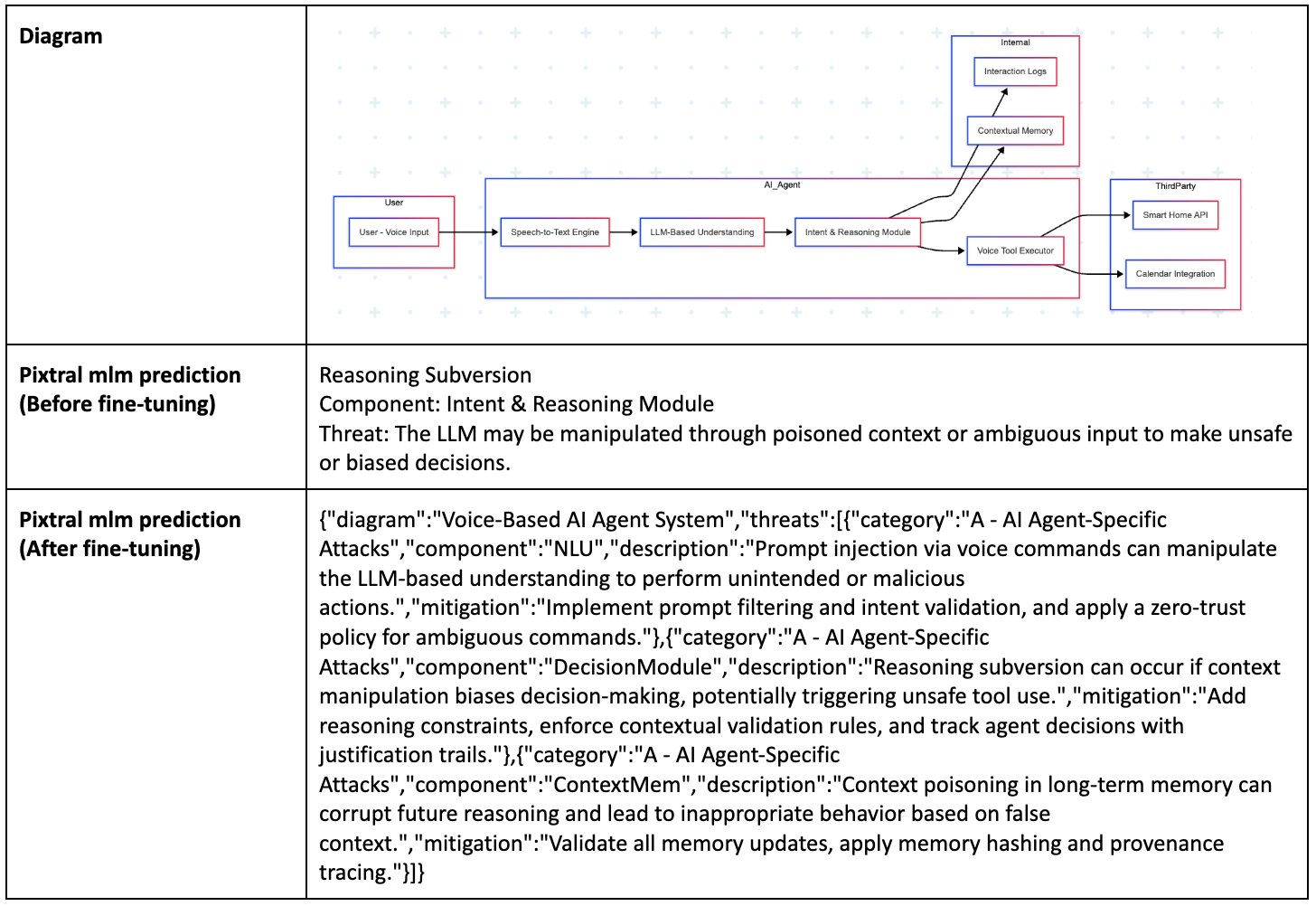}
\DeclareGraphicsExtensions.
\caption{The prediction results of Pixtral-Vision VLM.}
\label{pixtral-vision-result}
\end{figure}

Figure~\ref{llama-vistion-result} and Figure~\ref{pixtral-vision-result} present the predictions generated by two vision-language models, Llama-Vision and Pixtral-Vision~\cite{vistion-language-model-comparison}, before and after fine-tuning for the given AI agent-based system architectures. For Llama-Vision, the model initially detects only a single vulnerability—prompt injection at the prompt processor—without recognizing additional attack surfaces or providing detailed mitigations. After fine-tuning, it demonstrates substantial improvement by also identifying context poisoning in the reasoning core and unsafe tool invocation in the tool execution module. Furthermore, the model outputs context-aware mitigations, such as input sanitization, context integrity validation, and access control for API invocations. For Pixtral-Vision, the pre-fine-tuning prediction is similarly narrow, flagging only reasoning subversion at the intent and reasoning module. After fine-tuning, Pixtral-Vision expands its coverage to detect prompt injection in the NLU module, context poisoning in the contextual memory, and provides detailed, actionable mitigations—including prompt filtering with zero-trust validation, reasoning constraints with justification trails, and memory hashing with provenance tracking~\cite{llm-attacks}. Taken together, these results demonstrate that fine-tuning significantly enhances the capabilities of VLMs to perform structured ASTRIDE-based threat modeling. Both models show improved coverage of AI-agent–specific threats, better interpretation of architectural components (e.g., reasoning core, memory store, external APIs), and stronger alignment with expert-annotated threat data. This highlights the effectiveness of fine-tuning in enabling multimodal models to generate context-aware, explainable, and actionable security insights for complex agentic AI systems.

\subsection{Evaluation of OpenAI-gpt-oss Reasoning LLM} 

\begin{figure}[H]
\centering{}
\includegraphics[width=5.3in]{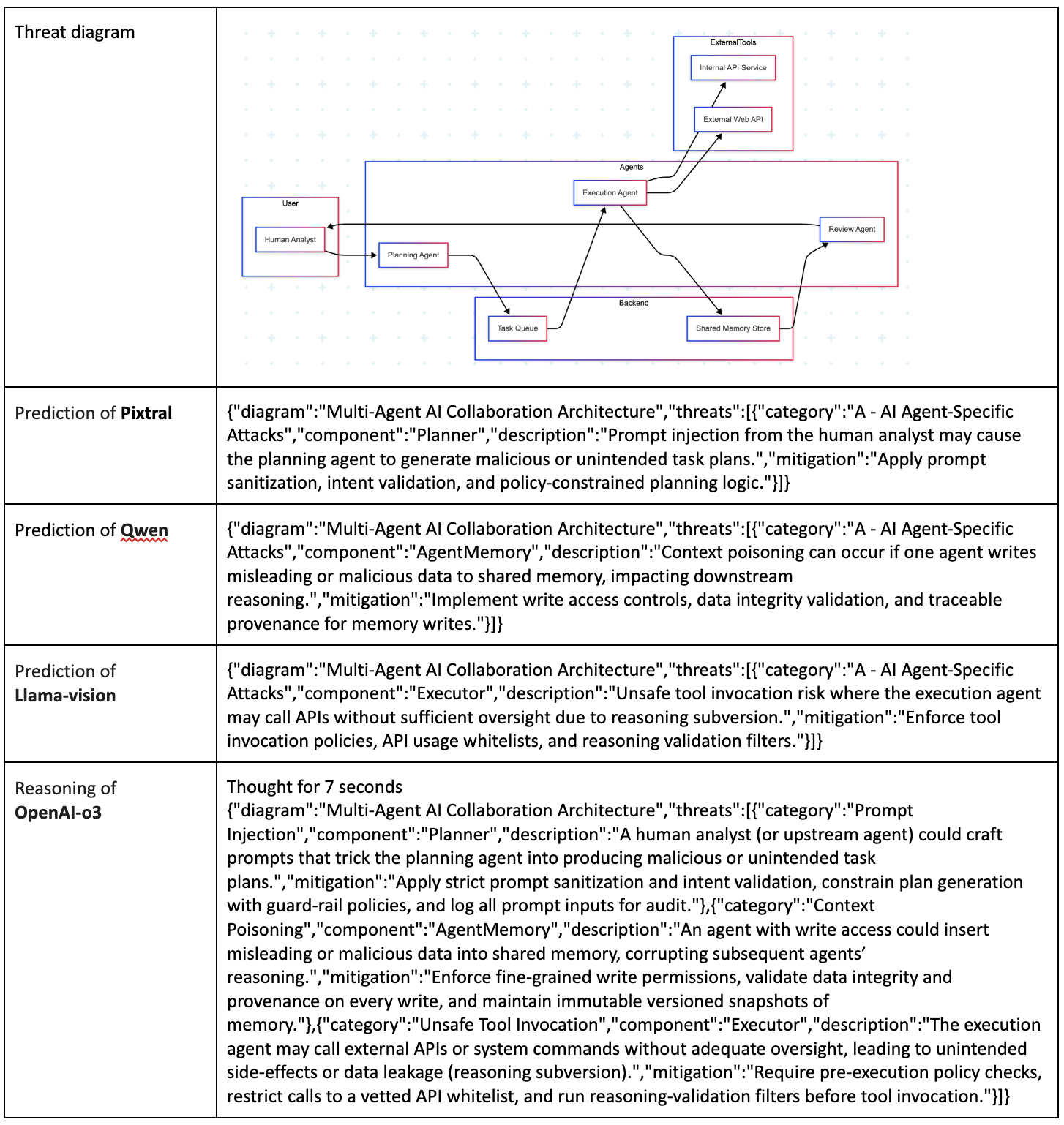}
\DeclareGraphicsExtensions.
\caption{Reasoning made by OpenAI-gpt-oss LLM.}
\label{gpt-oss-result}
\end{figure}

In this evaluation, we assessed the reasoning performance of the OpenAI-gpt-oss Reasoning LLM by comparing the individual predictions from the VLMs with the final prediction generated by OpenAI-gpt-oss. Figure~\ref{gpt-oss-result} illustrates the predictions made by different VLMs for a threat modeling diagram, along with the final reasoning provided by OpenAI-gpt-oss. The results highlight OpenAI-gpt-oss’s ability to synthesize and analyze the outputs from multiple VLMs, demonstrating its effectiveness and reliability in producing accurate final diagnoses. This evaluation underscores the reasoning LLM’s role in refining and enhancing prediction precision, further validating its integration into the Astride framework.

\section{Conclusions and Future Work}

In this paper, we introduce ASTRIDE, a novel, fully automated threat modeling platform tailored for AI agent-based systems. ASTRIDE extends the traditional STRIDE framework by introducing a new threat category \textit{A for AI Agent–Specific Attacks} to capture emerging security risks such as prompt injection, context poisoning, and unsafe tool invocation that are unique to agentic workflows. We fine-tuned a consortium of VLMs, including Pix2Struct, Qwen2-VL, and Llama-Vision, using a custom dataset of annotated agent architecture diagrams containing trust boundaries, component interactions, data flows, and ASTRIDE-labeled threat vectors. These fine-tuned models are capable of interpreting complex visual representations and generating structured, component-level threat predictions.
To enhance the reliability and transparency of the automation workflow, we employed LLM Agents to orchestrate secure and auditable interactions between the VLM consortium and the reasoning LLM. All VLMs were fine-tuned using the Unsloth library with QLoRA-based quantization, enabling resource-efficient deployment on consumer-grade hardware without sacrificing performance. Experimental evaluations using both synthetic and real-world architecture diagrams demonstrate that ASTRIDE significantly improves the accuracy of threat identification, reduces the dependence on human experts, and offers a scalable and explainable alternative to traditional threat modeling practices. For future work, we plan to expand the VLM ensemble with newer open-source multimodal models to further increase prediction performance and robustness.



\bibliographystyle{elsarticle-num}
\bibliography{reference}

\end{document}